\documentclass[10pt,twocolumn,letterpaper]{article}

\usepackage{cvpr}      %
\usepackage{algorithm}
\usepackage{algpseudocode}
\usepackage{multirow}
\usepackage{multicol}
\usepackage{colortbl}
\usepackage{pgfplots}
\usepackage{pgfplotstable}
\usepackage{tikz}
\usepackage{bm,bbm}
\usepackage{soul}
\usepackage{overpic}
\pgfplotsset{compat=newest} 
\DeclareMathOperator*{\argmax}{arg\,max}
\DeclareMathOperator*{\argmin}{arg\,min}
\usepackage{pifont}%
\newcommand{\xmark}{\ding{55}}%

\definecolor{cvprblue}{rgb}{0.21,0.49,0.74}
\usepackage[pagebackref,breaklinks,colorlinks,allcolors=cvprblue]{hyperref}
\usepackage[capitalize]{cleveref}
\def\paperID{8793} %
\def\confName{CVPR}
\def\confYear{2025}

\title{Condensing Action Segmentation Datasets via Generative Network Inversion}

\author{Guodong Ding, Rongyu Chen and Angela Yao\\
National University of Singapore\\
{\tt\small \{dinggd, rchen, ayao\}@comp.nus.edu.sg}
}

\begin{document}
\maketitle
\begin{abstract}

This work presents the first condensation approach for procedural video datasets used in temporal action segmentation. We propose a condensation framework that leverages generative prior learned from the dataset and network inversion to condense data into compact latent codes with significant storage reduced across temporal and channel aspects. Orthogonally, we propose sampling diverse and representative action sequences to minimize video-wise redundancy. Our evaluation on standard benchmarks demonstrates consistent effectiveness in condensing TAS datasets and achieving competitive performances. Specifically, on the Breakfast dataset, our approach reduces storage by over 500$\times$ while retaining 83\% of the performance compared to training with the full dataset. Furthermore, when applied to a downstream incremental learning task, it yields superior performance compared to the state-of-the-art.
\end{abstract}

\section{Introduction}\label{sec:intro}

The effectiveness of data-driven deep neural network models hinges on training with large and diverse datasets. For instance, ImageNet~\cite{deng2009imagenet} comprises over 14 million natural images for image classification, and Common Crawl~\cite{commoncrawl} provides over 5 billion website pages to foster natural language processing tasks. However, the storage, processing, and training costs associated with massive datasets pose significant challenges, especially as data volumes continue to grow. This need for efficiency has driven interest in \textit{dataset condensation}~\cite{wang2018dataset}, a technique aimed at compressing large datasets into smaller, information-rich subsets. Such techniques have potential applications in downstream areas, including incremental learning and federated learning.

The core objective of dataset condensation is to learn a small synthetic dataset from an original large-scale dataset.  Ideally, models trained on the synthetic data should perform comparably to those trained on the original. %
Significant efforts since then~\cite{cazenavette2022dataset,cui2023scaling,liu2022dataset,wang2022cafe,zhao2023dataset} have been dedicated to dataset condensation for image data.
Video, however, remains under-explored. {The recent work}~\cite{wang2024dancing} condenses static and dynamic information from videos for action recognition in a two-stage framework. %
First, a synthetic ``image'' aggregates the static visual information, while a 
dynamic memory block is learned to capture and supplement the motion and dynamics. 

\begin{figure}[t]
    \centering
    \hspace{-0.7em}
\includegraphics{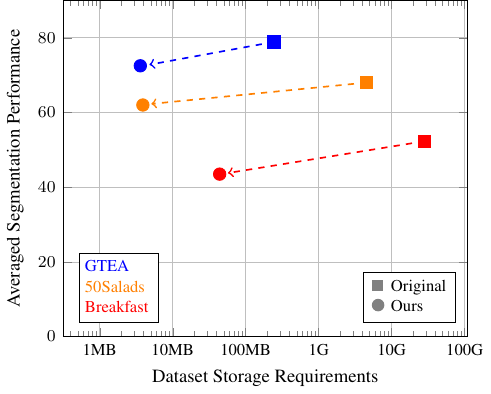}%
    \caption{Comparison of action segmentation performance with dataset storage across common action segmentation benchmarks at different scales. Our method effectively reduces dataset storage while retaining competitive performance to the original setup. %
    }
    \label{fig:teaser}
    \vspace{-0.2cm}
\end{figure}

This work aims to develop an effective dataset condensation approach for temporal action segmentation (TAS)~\cite{ding2023temporal}. TAS is a task that divides videos into segments by assigning labels on a per-frame basis to capture the sequence and duration of distinct actions. 
The videos are long, often spanning several minutes, creating significant challenges in raw data storage and processing.  Dataset condensation for TAS presents unique challenges that are not typically encountered for action recognition datasets.
First, TAS requires frame-level predictions rather than assigning a single label to an entire video or segment, as in action recognition. Consequently, dataset condensation for TAS must be capable of restoring the actual temporal resolutions of segments. Directly adapting the existing video condensation approach proposed by~\cite{wang2024dancing} for condensing TAS datasets is non-trivial.~\cite{wang2024dancing} uses a fixed-length frame processing, %
which cannot handle varying length segments while being able to restore the original temporal resolution. Additionally, the network in~\cite{wang2024dancing} is designed to work with RGB image sequences, while TAS typically uses pre-computed frame features~\cite{farha2019ms,yi2021asformer,liu2023diffusion}.
Second, the action ordering in the video sequences is not rigidly fixed, but there is still a degree of structure or dependency that governs how actions unfold. Some actions can occur flexibly in their sequence, while others must follow a specific progression. As a result, redundant action sequences can emerge, where identical patterns in action order are presented multiple times. This sequence redundancy is not present in AR datasets.

This work presents the first study on condensing TAS datasets. We propose a framework that condenses the dataset with a generative model into a set of latent codes. The generative model is first trained on the entire dataset to learn the prior. Then, through network inversion, we optimize a set of latent codes as the condensed data, minimizing the error between the generated and original frame features. 

Specifically, we use the TCA model proposed by~\cite{ding2024coherent} as our generative model. TCA is a conditional VAE reconstructing frame features while accounting for temporal dynamics.  It is well-suited for our task because it compresses and restores the temporal resolution of action segments through its coherence variable. TCA also condenses the feature dimension, as the latent code is more compact than the original feature. 
As a result, the storage requirement for a single action segment can be significantly reduced to the segment's latent code. This corresponds to the first challenge of TAS dataset condensation.

To address the second challenge of sequence redundancy, we propose a diversity-based sampling strategy grounded in the edit distance criterion. This approach iteratively selects the candidate sequence that maximizes the diversity within the chosen set. Empirically, we observed that TAS models achieve comparable performance with only half of the sequences. Our approach greatly reduces the storage while achieving comparable segmentation performance to the original dataset as shown in~\cref{fig:teaser}.

\textbf{Contributions.} Our contributions are summarized as follows:
\textbf{(1)} This work is the first to investigate and propose an effective dataset condensation approach for the temporal action segmentation task.
\textbf{(2)} We propose a generative condensation framework that first learns the generative prior on the dataset and then leverages the network inversion to condense data into compact latent codes. In addition, we propose a sampling strategy to further reduce the storage requirements based on sequence diversity.  
\textbf{(3)} Our approach effectively condenses TAS datasets of varying scales while consistently yielding comparable segmentation performances. Our approach outperforms the state-of-the-art by 10.7\% on Breakfast under an incremental setup.

\section{Related Work}\label{sec:related}
\textbf{Temporal Action Segmentation.}
Various approaches have been proposed to tackle the temporal action segmentation task~\cite{ding2023temporal}. Fully supervised methods require dense annotations for each video frame~\cite{farha2019ms, yi2021asformer}. In contrast, semi-supervised methods~\cite{singhania2022iterative, ding2022leveraging} only necessitate dense labels for a subset of the videos, leaving the rest unlabeled. There are also weaker forms of supervision, such as action transcripts~\cite{kuehne2017weakly}, action sets~\cite{richard2018action, li2020set, fayyaz2020sct}, timestamps~\cite{li2021temporal, rahaman2022generalized}, and activity labels~\cite{ding2022temporal}. Additionally, some approaches~\cite{sarfraz2021temporally, sener2018unsupervised, kukleva2019unsupervised} operate in an unsupervised setting without relying on any action labels.
Recent emerging directions include learning TAS incrementally~\cite{ding2024coherent} where procedural activities are learned sequentially and in an online~\cite{zhong2024onlinetas,shen2024progress} fashion. 
However, the storage burden of these video datasets remains a pressing issue as these videos are long and with rich redundancy. %
Our work is the first to study the dataset condensation of temporal action segmentation datasets from the perspective of efficient data storage.

\noindent \textbf{Dataset condensation.}
Dataset condensation (DC) is first formally introduced in~\cite{wang2018dataset}. The target is to learn a small set of synthetic data from an original large-scale dataset so that models trained on the synthetic dataset can perform comparably to those trained on the original. 
Based on the bilevel optimization formulation,
varying techniques are leveraged to alleviate optimization difficulty including group optimization~\cite{zhou2022dataset}, the Neural Tangents Kernel (NTK)~\cite{nguyen2020dataset,nguyen2021dataset}, empirical Neural Network Gaussian Process (NNGP)~\cite{loo2022efficient}.
Instead of matching model performance, \cite{zhao2020dataset,zhao2021dataset,kim2022dataset,cazenavette2022dataset,cui2023scaling} aim to indirectly achieve this by matching model parameters trained on original and condensed datasets. %
Another prominent line of work bypassing the bilevel optimization directly matches the distribution of original and condensed synthetic data~\cite{wang2022cafe,zhao2023dataset}, known as distribution matching.
Differently, \cite{cazenavette2023generalizing,zhang2023dataset} introduced the generative model and synthesized in its latent space.
Yet there is few work that extends this to the video domain. The work most related to ours is the dataset condensation for action recognition~\cite{wang2024dancing}. They disentangle the
static and dynamic information in the video to minimize temporal redundancy. 
However, their method is not applicable to temporal action segmentation as they can not account for varying temporal resolution of actions and video redundancy. Thus, we are motivated to propose a generative framework to tackle the procedure video condensation problem for temporal action segmentation.

\section{Dataset Condensation for TAS}\label{sec:approach}
\subsection{Preliminaries}

\begin{figure*}[htb]
    \centering
    \includegraphics[width=\textwidth]{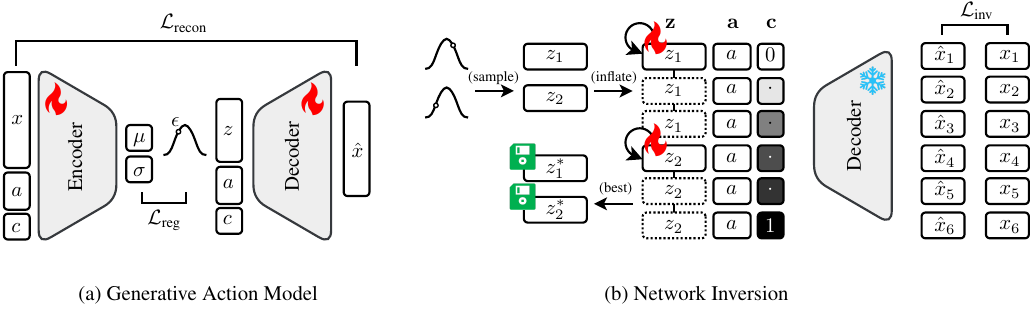}
    \caption{Generative Feature and Temporal Condensation Framework.  (a) The generative action model is a conditional VAE that is trained to reconstruct the input frames conditioned on the action class label and a coherence variable. (b) The network inversion aims to optimize between decoded and original segments. Randomly sampled latent codes $z_1$ and $z_2$ are first inflated over time to the segment length, then concatenated with the action label and coherence variable for decoding. During the optimization, only the latent codes get updated while the decoder always stays fixed. These optimized latent codes $z_1^*$ and $z_2^*$ are stored as the condensed representation of the original segment. \raisebox{-0.3ex}{\includegraphics[height=1em]{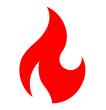}} indicates parameter updates during learning, while the \raisebox{-0.3ex}{\includegraphics[height=1em]{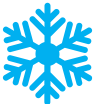}} indicates that the parameter is kept frozen.}\label{fig:dd}
    \vspace{-0.5em}
    \end{figure*}

\textbf{Temporal Action Segmentation (TAS).} Temporal action segmentation is a task that segments untrimmed procedural videos into contiguous and non-overlapping actions. For example, given a video $\mathbf{X} = \{x^1,..., x^T\}$ of $T$ frames long, a segmentation model outputs $N$ continuous action segments consecutive in time: 
\begin{equation}\label{eq:segment}
    \mathbf{s}_{1:N} = (s_1,s_2,...,s_N), \quad \text{where } 
    s_n = (a_n, t_n, \ell_n),
\end{equation}
\begin{equation}
   \text{s.t.} \qquad t_{n+1} = t_n+\ell_n. \nonumber
\end{equation}
$s_n$ is a segment of length of $\ell_n$, with action class label $a_n \in \mathcal{A}$ from $A$ predefined categories. The $t_n$ denotes the starting timestamp of segment $s_n$. %
Alternatively, most existing works~\cite{farha2019ms,yi2021asformer,liu2023diffusion} formulate it as a frame-wise classification task ($\mathcal{L}_{\text{cls}}$) and encourage the continuity of action segments with a smoothing term ($\mathcal{L}_{\text{sm}}$) with the learning objective written as:
\begin{equation}\label{eq:iltas}
    \mathcal{L}_{\text{tas}} = \mathcal{L}_{\text{cls}}(x,y) + \lambda \cdot \mathcal{L}_{\text{sm}}(x,y),
\end{equation}
where $y\in\mathcal{A}$ is the action label and $\lambda$ a trade-off parameter. %

\noindent \textbf{Dataset Condensation (DC).} 
Let $\mathcal{R} = \{\mathbf{X}_r, \mathbf{Y}_r\}$ represent a real image dataset, where $\mathbf{X}_r \in \mathbb{R}^{n_r \times d}$ denotes the set of training samples and $\mathbf{Y}_r \in \mathbb{R}^{n_r \times c}$ corresponds to their associated labels. Here, $n_r$ denotes the number of original samples while $d$ and $c$ represent the dimensionality of the input features and output labels, respectively. The objective of dataset condensation is to construct a synthetic dataset of $n_s$ samples, \ie $\mathcal{S} = \{\mathbf{X}_s, \mathbf{Y}_s\}$, where $\mathbf{X}_s \in \mathbb{R}^{n_s \times d}$ and $\mathbf{Y}_s \in \mathbb{R}^{n_s \times c}$, with a considerably reduced size compared to the real dataset, such that $n_s \ll n_r$. 

\subsection{Task Formulation}
Given an original TAS dataset with $n_r$ training videos, represented as $\mathcal{R} = \{(\mathbf{X}_i, \mathbf{Y}_i)\}_{i=1}^{n_r}$. Each video representation $\mathbf{X}\in\mathbb{R}^{T\times D}$ and its corresponding action label $\mathbf{Y}\in\mathbb{R}^{T\times A}$ have the same temporal length $T$. Here, $D$ and $A$ denote the dimensions of frame feature space ($x\in\mathbb{R}^D$) and the action space ($y\in\mathbb{R}^A$), respectively. The goal of dataset condensation is to create a compact subset $\mathcal{S}=\{\hat{\mathbf{X}}_i, \hat{\mathbf{Y}}_i\}_{i=1}^{n_s}$, where $\hat{\mathbf{X}}\in\mathbb{R}^{T'\times d}$ represents a condensed version of original video. The size of $\mathcal{S}$ is expected to be significantly smaller than that of the original dataset $\mathcal{R}$. 

This objective can be achieved through two levels of dataset condensation: (1) Sample compression: reducing the dimensionality of each video such that $T'\times d \ll T\times D$ and (2) Sample Reduction: reducing the total number of samples, \ie, $n_s \ll n_r$.
In light of this, our condensation framework implements reductions at both levels: for sample compression, we propose a generative feature and temporal condensation technique using network inversion~(\cref{subsec:featandtemp}). For sample reduction, we use a diversity-based sampling strategy~(\cref{subsec:seqdiv}).

\subsection{Generative Feature \& Temporal Condensation}\label{subsec:featandtemp}

Generative models are compact yet flexible, able to produce outputs of various lengths, making them ideal for condensing TAS datasets. The condensation process involves two main stages. In the first stage, a generative model is trained to represent action segments. %
In the second stage, a network inversion process is applied to optimize the latent codes, ensuring the condensed dataset captures an optimal representation of the original action segments.

\noindent \textbf{Generative Action Model.} 
We choose the Temporally Coherent Action (TCA) model proposed in~\cite{ding2024coherent} as our generative action model. TCA is essentially a compact, two-layer MLP VAE trained to reconstruct frame features. Specifically, the encoder in the TCA model takes three inputs: frame feature $x$, action label $a$, and a coherence variable $c$. The variable $c$ is mathematically defined as the relative position of the frame within its segment:
\begin{equation}\label{eq:c}
    c_i = (i-1)/(\ell-1), \quad \text{and}\quad c_i \in [0,1].
\end{equation}

The VAE's encoder maps these inputs in a latent space while the decoder reconstructs the frame feature $\hat{x}$. We denote the encoder and decoder as $\mathbf{E}(x,a,c)=q_\phi(z|x,a,c)$ and $\mathbf{D}(z,a,c)= p_\theta(x|z,a,c)$, respectively. The TCA model is trained on the entire dataset's video frames with a reconstruction loss and a KL divergence regularizer:
\begin{equation}\label{eq:cvae}
        \mathcal{L}_{\text{TCA}} = \underbrace{\mathbb{E}_z \log p_{\theta}(x|z,a,c)}_{\mathcal{L}_{\text{recon}}} - \underbrace{\text{D}_{\text{KL}}(q_{\phi}(z|x,a,c)||p(z))}_{\mathcal{L}_{\text{reg}}}.
\end{equation} 
An overview of the generative action model is depicted in~\cref{fig:dd}\hyperref[fig:dd]{(a)}. 
In this way, a segment of $\mathbb{R}^{\ell \times D}$ can be efficiently compressed to a latent distribution characterized by the mean and standard deviation $(\mu, \sigma)\in\mathbb{R}^d$, $d$ is the dimension of the latent space. An advantage of using a generative model for condensing TAS videos is the ability to restore the original resolution. 
Such a model can generate segments of any specified length $\ell$, producing each frame $\hat{x}$ by decoding a randomly sampled latent code $z$ as follows:
\begin{equation}\label{eq:gen}
    \hat{x}_i = p_\theta (x|z, a, c_i), \quad \text{and } i \in [1, ..., \ell].
\end{equation}

\noindent To ensure temporal continuity of generated features, \cite{ding2024coherent} suggests a fixed latent code $z$ is applied across all frames within the same segment.

\noindent \textbf{Network Inversion.} 
The above model learns inherent action priors from the video dataset, enabling it to generate segments that reflect realistic actions. However, segments decoded from \textit{randomly sampled} latent codes can still deviate significantly from real data. To limit the deviations, we propose using network inversion.
Neural network inversion is the process of determining a neural network input when given the corresponding output.  Formally, given a neural network $f:\mathbb{R}^n \rightarrow \mathbb{R}^m$ that maps an input $x\in \mathbb{R}^n$ to an output $y\in \mathbb{R}^m$, where $y=f(x)$. %
Mathematically, given an output $y$, the objective of neural network inversion is to find an input $x^*$ such that:
\begin{equation}
    x^*=f^{-1}(y),
\end{equation}
where $f^{-1}$ represents an approximate or exact inverse of function $f$. 
Since neural networks are generally not invertible, the problem can be posed as an optimization problem:
\begin{equation}\label{eq:opt}
    x^* = \argmin_{x\in\mathbb{R}^d} \mathcal{C}(f(x),y),
\end{equation}
where $\mathcal{C}(\cdot, \cdot)$ is a cost function. Note that during the inversion, both $f(\cdot)$ and $y$ remain fixed, while only $x$ is updated. 

As our generation depicted in~\cref{eq:gen} is at the segment level, the inversion objective from~\cref{eq:opt} becomes:
\begin{equation}\label{eq:ouropt}
    z^* = \argmin_{z\in\mathbb{R}^d} \underbrace{||\mathbf{D}(\mathbf{z},\mathbf{a},\mathbf{c})-\mathbf{x}||_2^2}_{\mathcal{L}_\text{inv}},
\end{equation}
where $\mathbf{z}=z\otimes\mathbf{1}_\ell$, $\mathbf{a}=a\otimes\mathbf{1}_\ell$, and $\mathbf{c}=[c_1,\dots,c_\ell]$. $\mathbf{1}_\ell$ is a vector of ones of length $\ell$. 
We choose the $\ell_2$ norm as the cost function in the inversion loss $\mathcal{L}_\text{inv}$ to align with the reconstruction term $\mathcal{L}_\text{recon}$ in the generative model training (as shown in~\cref{eq:cvae}). Upon performing the inversion, the optimized latent code $z^*\in \mathbb{R}^{d}$ is stored for each segment.

\noindent \textbf{Instances per Segment.} In TAS datasets, action segments can be particularly long, where a single global latent code may not suffice to restore the full complexity and temporal dynamics of the entire segment. This limitation highlights the need for finer-grained approximations. To address this, we introduce the concept of \textit{instances per segment}, which divides each segment into smaller, finer-grained instances for more precise network inversion. This is akin to the instances per class commonly used in existing dataset condensation works~\cite{yu2023dataset}. We evenly split segments into smaller instances to enable inversions at local scales. 

Specifically, during the inversion step, for a given segment, we first initialize a set of $K$ random codes $\{z_k\}_{k=1}^K$. These codes are evenly inflated over time to match the actual length of the segment, yielding the vector $\mathbf{z}=[z_1\otimes \mathbf{1}_{\ell_1}, ..., z_K \otimes \mathbf{1}_{\ell_K}]$, where $\ell_k =\frac{\ell}{K}$. The vector $\mathbf{z}$ is concatenated with the action label $\mathbf{a}$ and the coherence variable $\mathbf{c}$, and the combined input is fed into the decoder for network inversion as defined in~\cref{eq:ouropt}.
After performing the inversion, we store the set of optimal latent codes $\{z_k^*\}_{k=1}^K$ as the condensed representation of the segment's features. 
An illustrative depiction of the inversion process for a segment of length $\ell=6$ with $K=2$ instances per segment is shown in~\cref{fig:dd}\hyperref[fig:dd]{(b)}. Initially, two latent codes $z_1, z_2$ are randomly sampled and expanded temporally to generate the segment $\{\hat{x}\}$ through the decoder. The decoder remains fixed while only the latent codes are optimized. Once optimized, the final $z_1^*,z_2^*$ are stored as the condensed segment.  %

The proposed framework simultaneously condenses feature and temporal dimensions and reduces the storage requirement for each segment. %
Specifically, a segment $\mathbf{x}\in\mathbb{R}^{\ell\times D}$ can be efficiently condensed into latent codes $\mathbf{z} \in \mathbb{R}^{K\times d}$, with the condensation factor given by $\frac{\ell\cdot D}{K\cdot d}$. 
At the video level, our framework condenses the original  $\mathbf{X}\in\mathbb{R}^{T\times D}$ into a reduced representation $\hat{\mathbf{X}}^*\in\mathbb{R}^{KN\times d}$, where $N$ denotes the number of segments in the video, which is substantially smaller than the original video length $T$. 

\subsection{Diverse Sequence Sampling}~\label{subsec:seqdiv}
The condensation process described above occurs at the sample level, reducing both feature and temporal dimensions. To account for sample redundancy, we introduce a %
diversity-based pruning strategy.
Our intuition is that the selected sequences, taken collectively, should capture the maximum diversity of action ordering within the dataset. This ensures that the pruned set retains the broadest range of unique temporal patterns and action variations. Edit distance measures the minimum operations needed to transform one sequence into another, making it suitable for quantifying sequence diversity. 
Given two action sequences $\mathbf{s}_i$ and $\mathbf{s}_j$, we quantify the diversity with the normalized edit distance between them:
\begin{equation}
    \text{Edit}(\mathbf{s}_i,\mathbf{s}_j) = \frac{e[|\mathbf{s}_i|, |\mathbf{s}_j|]}{\max (|\mathbf{s}_i|,|\mathbf{s}_j|)}, \quad \text{and } 
\end{equation}
\noindent\scalebox{0.9}{\parbox{\linewidth}{%
\begin{equation}
e[m,n] =
 \begin{cases}
\max(m,n),&\!\!\min(m,n)\!=\!0\\
 \begin{aligned}
\min(&e[m\!-\!1,n]\!+\!1,e[m,n\!-\!1]\!+\!1,\\&e[m\!-\!1,n\!-\!1]\!+\!\mathbbm{1}(\mathbf{s}_i^m\!\neq \!\mathbf{s}_j^n))
 \end{aligned}, &  \!\!\text{otherwise.}\nonumber
\end{cases}
\label{eq:edit}
\end{equation}
}}
where $m,n$ denote the action index within two comparing sequences, respectively. $\mathbbm{1}(\cdot)$ is an indicator function. 

We then apply a furthest point sampling strategy, commonly used in point clouds~\cite{qi2017pointnet++}, to progressively select sequence $\mathbf{s}^*$ that maximizes the diversity until the desired set cardinality is reached. Specifically:
\begin{equation}\label{eq:div}
    \mathbf{s}^* = \argmax_{\mathbf{s}_i\in \mathcal{D}\setminus S}\min_{\mathbf{s}_j \in  \mathcal{S}} \text{Edit}(\mathbf{s}_i, \mathbf{s}_j),
\end{equation}
where $\mathcal{D}$ is the original dataset and $\mathcal{S}$ the selected set, and $|\mathcal{S}| =\gamma|\mathcal{D}|$.
We empirically set the size of the sampled to half of the original dataset, \ie, $\gamma=0.5$. This yields an extra $\sim\!50$\% reduction in the storage of latent codes.

\subsection{Decoding for TAS}
Neural networks are sensitive to input resolution, and training a TAS model on low-resolution or condensed input can lead to suboptimal performance. Therefore, restoring the original resolution of input data is essential for the segmentation model to learn effectively. Different than the random generation in~\cite{ding2024coherent}, we restore the action segments with their respective latent codes $\{z_k^*\}$, action labels $a$ and length $\ell$ (coherence variable $c$), with the decoder $\mathbf{D}$ as follows:
\begin{equation}
    \hat{\mathbf{x}}^* = \mathbf{D}(\mathbf{z}^*, \mathbf{a}, \mathbf{c}),
\end{equation}
where $\mathbf{z}^*=[z_1^*\otimes \mathbf{1}_{\ell_1}, ..., z_K^* \otimes \mathbf{1}_{\ell_K}]$.
These restored segments $\hat{\mathbf{x}}^*$ are then concatenated in time to form videos $\hat{\mathbf{X}}^*$, and their temporal order follows the symbolic sequence stored in the pruned set $S$. 
Hence, the training objective in~\cref{eq:iltas} of the segmentation model becomes:
\begin{equation}
    \mathcal{L}_{\text{tas}} = \mathcal{L}_{\text{cls}}(\hat{x}^*,y) + \lambda \cdot \mathcal{L}_{\text{sm}}(\hat{x}^*,y),
\end{equation}
Details of the loss terms are given in the Supplementary.

\section{Experiments}\label{sec:exp}
\subsection{Datasets and Evaluation}

\textbf{Datasets.} We evaluate our approach on three common TAS benchmarks that vary in storage scales. 
\textbf{GTEA}~\cite{fathi2011learning} contains 28 videos of seven kitchen activities composing 11 different actions. \textbf{50Salads}~\cite{stein2013combining} has 50 videos with 19 action classes. 
\textbf{Breakfast}~\cite{kuehne2014language} dataset comprises 1,712 undirected breakfast preparation videos. There are 10 activities and a total of 48 action classes; each video features 5 to 14 actions. 
In terms of \textbf{storage}, the three datasets are at three scales: GTEA is the smallest at 245 MB, 50Salads is in the middle at 4.5 GB, and Breakfast is the largest at 28 GB.
For all datasets, we use the I3D~\cite{carreira2017quo} feature representations and evaluate with the standard splits. Although I3D initially compresses frames by transforming RGB data into feature space, the original temporal resolution remains.

\noindent \textbf{Evaluation Measures.} 
TAS is evaluated using three metrics: frame-wise accuracy (Acc), segment-wise edit score (Edit), and F1 score with varying overlap thresholds of 10\%, 25\%, and 50\%. In addition to these conventional TAS metrics, we also report the storage size to highlight the level of dataset condensation. %

\subsection{Implementation}
\textbf{Generative Network Inversion.} We use the TCA~\cite{ding2024coherent} as our generative model, and follow their implementation as a two-layer MLP for both encoder and decoder with the latent size $d=256$. On each dataset, we train the model for $7.5K$ epochs with a learning rate of $1e^{-3}$. 
For the network inversion, we optimize~\cref{eq:ouropt} for $10K$ iterations to obtain the optimal latent codes $z^*$. In all our experiments, unless otherwise specified, we set the number of instances per segment $K=8$ and the sequence sampling ratio $\gamma=0.5$. 

\setlength{\fboxsep}{0pt}

\begin{table*}[!t]
\centering
\resizebox{\textwidth}{!}{
\begin{tabular}{@{}l@{\hskip 5pt}c@{\hskip 6pt}c@{\hskip 6pt}c@{\hskip 5pt}r@{\hskip 6pt}c@{\hskip 6pt}c@{\hskip 6pt}c@{\hskip 5pt}r@{\hskip 6pt}c@{\hskip 6pt}c@{\hskip 6pt}c@{\hskip 5pt}r@{}}
\toprule
 & \multicolumn{4}{c}{GTEA~\cite{fathi2011learning}} & \multicolumn{4}{c}{50Salads~\cite{stein2013combining}} & \multicolumn{4}{c}{Breakfast~\cite{kuehne2014language}} \\ \cmidrule(lr){2-5}\cmidrule(r){6-9}\cmidrule(r){10-13}
 & Acc & Edit & F1@\{10, 25, 50\} & Storage & Acc & Edit & F1@\{10, 25, 50\} & Storage & Acc & Edit & F1@\{10, 25, 50\} & Storage \\
 \midrule
&\multicolumn{12}{c}{MS-TCN~\cite{farha2019ms}}\\ 
 \midrule
Original & 79.0 & 76.3 & 85.8 / 83.4 / 69.8 & \textcolor{red}{245 MB} & {80.6} & 63.1 & 69.9 / 67.4 / 59.0 & \textcolor{red}{4.5 GB} & 67.2 & 60.6 & 50.5 / 46.3 / 36.8 & \textcolor{red}{28 GB} \\\midrule
Mean & 71.2 &\textbf{ 73.3} & 77.1 / 73.7 / 59.4 & \textcolor{orange}{7.2 MB} & 69.0 & 42.7 & 50.0 / 46.1 / 37.4 & \textcolor{orange}{7.8 MB} & 47.6 & 31.8 & 27.8 / 23.3 / 15.6 & \textcolor{orange}{96 MB} \\
Coreset~\cite{welling2009herding} & 66.7 & 66.1& 72.4 / 68.9 / 53.2&\textcolor{orange}{7.2 MB} &61.7&43.3& 49.9 / 46.3 / 35.4 &\textcolor{orange}{7.8 MB}& 49.7 & 36.8& 32.3 / 27.5 / 19.3 &\textcolor{orange}{96 MB}\\
TCA~\cite{ding2024coherent}&60.9&54.1&59.2 / 55.3 / 39.3 & \multicolumn{1}{c}{-} & 56.4 & 33.6&39.8 / 35.8 / 25.9  &\multicolumn{1}{c}{-}&34.2&20.7& 17.9 / 13.8 / 8.4 & \multicolumn{1}{c}{-} \\
Encoded & 70.4 & 65.5 & 72.2 / 68.8 / 52.1 & \textcolor{blue}{3.6 MB} & 69.0 & 43.6 & 50.6 / 46.0 / 37.4  & \textcolor{blue}{3.9 MB} & 37.9 & 49.8 & 40.0 / 32.8 / 19.4 & \textcolor{blue}{44 MB}  \\ 
\rowcolor{gray!30}
Ours & \textbf{75.2} & 71.9 & \textbf{78.3} / \textbf{74.6} / \textbf{62.7} & \textcolor{blue}{3.6 MB} & \textbf{74.4} & \textbf{59.5} & \textbf{65.1 }/ \textbf{61.0} /\textbf{ 50.2 }& \textcolor{blue}{3.9 MB} & \textbf{55.5} & \textbf{45.6} & \textbf{46.7} / \textbf{41.1} / \textbf{28.7} & \textcolor{blue}{44 MB} \\ \midrule
Encoded$^\dagger$ & 70.5  & 72.7 &  77.1 /  73.7 / 59.8 & 30.5 MB & 72.1 &  58.2 & 63.2 / 60.0  / 49.3  & 564 MB & 43.4 & 53.2 & 45.8 / 37.4 / 22.8 & 3.4 GB \\ \rowcolor{gray!30}
Ours$^\dagger$ & \textbf{73.3} & \textbf{73.8} & \textbf{79.2} / \textbf{75.4} / \textbf{65.5} & 30.5 MB & \textbf{72.8} & \textbf{59.8} & \textbf{65.2 }/ \textbf{61.3} / \textbf{51.3} & 564 MB & \textbf{54.1} & \textbf{53.3} & \textbf{49.8} / \textbf{44.3} / \textbf{33.1} & 3.4 GB \\ \midrule
&\multicolumn{12}{c}{ASFormer~\cite{yi2021asformer}}\\ 
 \midrule
 Original & 79.7 & 84.6 & 90.1 / 88.8 / 79.2 & \textcolor{red}{245 MB} & 85.6 & 79.6 & 85.1 / 83.4 / 76.0 & \textcolor{red}{4.5 GB} & 73.5 & 75.0 & 76.0 / 70.6 / 57.4 & \textcolor{red}{28 GB} \\\midrule
Mean & 72.2 & 76.9 & 82.1 / 79.7 / 65.1 & \textcolor{orange}{7.2 MB} & 71.6 & 49.8 & 56.6 / 52.5 / 43.4 & \textcolor{orange}{7.8 MB} & 52.2 & 43.2 & 43.5 / 38.3 / 26.7 & \textcolor{orange}{96 MB} \\
Coreset~\cite{welling2009herding} & 71.0 &75.4 & 81.0 / 78.1 / 62.9 &\textcolor{orange}{7.2 MB} & 69.4 &46.8  & 56.6 / 52.9 / 39.6 &\textcolor{orange}{7.8 MB}& 52.0 &48.1& 48.3 / 42.4 / 29.7&\textcolor{orange}{96 MB}\\
TCA~\cite{ding2024coherent}&62.2 & 57.8& 63.0 / 57.4 / 39.9 & \multicolumn{1}{c}{-}& 66.8 & 44.0 & 52.2 / 47.3 / 36.6 &\multicolumn{1}{c}{-}&36.6&28.2& 26.3 / 22.1 / 14.3&\multicolumn{1}{c}{-} \\
Encoded & 69.2 & 70.2 & 73.3 / 67.3 / 49.8 & \textcolor{blue}{3.6 MB} &  71.2 & 45.4 & 55.0 / 50.4 / 40.2 & \textcolor{blue}{3.9 MB} & 37.6 & 53.6 & 50.7 / 41.3 / 24.0 & \textcolor{blue}{44 MB}  \\ 
\rowcolor{gray!30}
Ours & \textbf{77.9} & \textbf{82.7} & \textbf{86.4} / \textbf{84.5} / \textbf{70.4} & \textcolor{blue}{3.6 MB} & \textbf{81.2} & \textbf{68.9} & \textbf{77.0} / \textbf{73.8} / \textbf{64.7} & \textcolor{blue}{3.9 MB} & \textbf{59.8} & \textbf{48.8} & \textbf{54.1} / \textbf{47.7} / \textbf{34.1} & \textcolor{blue}{44 MB} \\ \midrule
Encoded$^\dagger$ & 74.0 & 78.1 & 83.1 / \textbf{79.6} / 67.3 & 30.5 MB & 75.6 & 60.1 & 67.7 / 64.2 / 53.5 & 564 MB & 45.7 & 54.8 & 52.6 / 43.3 / 25.2 & 3.4 GB \\ 
\rowcolor{gray!30}
Ours$^\dagger$ & \textbf{75.0} & \textbf{79.0} & \textbf{83.6} / 79.5 / \textbf{67.7} & 30.5 MB & \textbf{76.2} & \textbf{65.0} & \textbf{73.1} / \textbf{68.8} / \textbf{58.5} & 564 MB & \textbf{61.1} & \textbf{61.4} & \textbf{62.4} / \textbf{56.0} / \textbf{42.1} & 3.4 GB \\ 
\bottomrule
\end{tabular}}
\caption{Performance comparison on dataset condensation for TAS on three common benchmarks with different backbones. Storage sizes are highlighted in colors (\textcolor{red}{high}, \textcolor{orange}{medium}, \textcolor{blue}{low}). Our method remarkably reduces storage while retaining competitive performances across different datasets and model architectures. More details of the settings ($d$, $K$, and $\gamma$) for each method are provided in the Supplementary. %
}\label{tab:effectiveness}
\vspace{-0.5em}
\end{table*}

\noindent \textbf{{Segmentation Backbones.}} We evaluate the effectiveness of our dataset condensation framework with two popular TAS backbones, \ie, MSTCN~\cite{farha2019ms} and ASFormer~\cite{yi2021asformer}. The former is a convolution-based segmentation model, while the latter is based on transformer architectures. 
We train MSTCN with a learning rate of $5e^{-4}$ for 50 epochs and  $1e^{-4}$ for 30 epochs with ASFormer. 

\noindent \textbf{Baselines.} As the first work to address dataset condensation for TAS, we establish the following baselines for comparison. Recognizing that storage size is a key evaluation aspect of dataset condensation approaches, we vigorously implement the following with aligned storage sizes to ensure fair comparisons:

\noindent -- ``\textbf{Original}'' uses features of standard TAS datasets and no dataset condensation techniques are applied. 

\noindent -- ``\textbf{Mean}'' is a straightforward method that stores the average frame features of action segments as representatives. During TAS training, each average feature is repeated to match the segment length, creating a static \textit{boring} video~\cite{zhao2020dataset}. This method effectively reduces video length to the number of segments, \ie, $\mathbb{R}^{T\times D} \rightarrow \mathbb{R}^{N\times D}$.

\noindent -- ``\textbf{Coreset}'' utilizes the Herding~\cite{welling2009herding}  to identify the frame feature closest to the mean feature of the segment. The selected frames are then upsampled similarly to ``{Mean}'' to restore the original temporal resolution. Therefore, they have the same condensation ratio.

\noindent -- ``\textbf{TCA}''~\cite{ding2024coherent} is a baseline that follows its original implementation in which action segments are generated directly from random latent codes. This method does not require storage for latent codes, as they can be sampled on the fly during decoding.

\noindent -- ``\textbf{Encoded}'' is the closest to our setup, with the key difference being that, instead of using network inversion to obtain latent codes, it stores the mean of encoded segment frames. Specifically, $z_k = \text{mean}(\mu_1, ..., \mu_{\ell_k})$. This approach results in the same storage requirement as ours.

\noindent -- ``\textbf{Encoded$^\dagger$}'' refers to a setup similar to ``{Encoded}''  except for removing the sequence sampling and setting the number of instances per segment to the actual segment length, \ie, $K=\ell$, which creates a latent code for each individual frame. The approach condenses along the feature dimension rather than the temporal dimension.

\subsection{Effectiveness}
\Cref{tab:effectiveness} compares our approach to the baselines on three widely adopted TAS benchmarks.  As observed, approaches like ``Mean'' and ``Coreset'', which primarily condense from the temporal aspect, achieve similar performance across all datasets while maintaining an identical storage size. Note that in the best scenario, \textit{boring} videos generated by these approaches can account for up to 80\% performance of training with the ``Original''. This highlights the temporal redundancy present in videos.  
TCA~\cite{ding2024coherent} does not incur additional storage requirements for the latent code, yet it produces the lowest overall performance across all evaluation metrics on three datasets. Although the generated segments inherit the action priors learned from the dataset, it is still likely the decoded segments from randomly sampled latent codes may not align well with the original data. A segmentation model trained on these misaligned features may not generalize well to the real testing data.

By storing encoded mean features of segments from the encoder as latent codes and diverse sequence sampling, ``Encoded'' can manage to achieve segmentation performance comparable to the ``Mean'' baseline, while requiring only half the storage cost. %
The best performance is achieved by our approach, which adds a network inversion process on top of ``Encoded''. By imposing network inversion, a significant performance gain in segmental metrics is observed. For instance, on the 50Salads dataset, the average F1 score is boosted by a substantial 14.1\% (from 44.7\% to 58.8\%). This underscores the effectiveness of network inversion, as it adapts the latent codes to better reflect the actual data. 

Comparing across storage sizes, our approach also significantly outperforms its counterparts, ``Mean'' and ``Coreset'', while only requiring roughly half the storage burden -- 44 MB compared to 96 MB on the Breakfast dataset.

Our proposed condensation framework is independent of the segmentation model, making it compatible with different backbones. 
TAS performances in~\cref{tab:effectiveness} with two  segmentation backbones~\cite{farha2019ms,yi2021asformer} demonstrates consistent performance improvements over the baselines.  
\begin{table}[!tb]
\centering
    \begin{tabular}{lc@{\hskip 3pt}c@{\hskip 6pt}c@{\hskip 6pt}c@{\hskip 6pt}r}
        \toprule
         & Sampling & Acc & Edit & F1@\{10, 25, 50\} &Storage \\ \midrule
        \multirow{3}{*}{\rotatebox{90}{GTEA}} & \xmark & 76.3 & 74.8 & 80.0 / 78.0 / 61.7 & 7.2 MB\\
         & Random  & 74.0 & 69.3 & 76.2 / 72.8 / 60.4 & 3.6 MB\\
         & \cellcolor{gray!30}{Ours} & \cellcolor{gray!30}{75.2} & \cellcolor{gray!30}{71.9} & \cellcolor{gray!30}{78.3 / 74.6 / 62.7} & \cellcolor{gray!30}{3.6 MB}\\
        \midrule
        \multirow{3}{*}{\rotatebox{90}{50Salads}} &  \xmark & 75.3 & 60.0 & 66.2 / 62.6 / 49.9 & 7.8 MB\\
         &  Random & 71.9 & 58.9 & 62.3 / 58.4 / 49.5&3.9 MB\\
         & \cellcolor{gray!30}{Ours} & \cellcolor{gray!30}{74.4} &\cellcolor{gray!30}{59.5} & \cellcolor{gray!30}{65.1 / 61.0 / 50.2} & \cellcolor{gray!30}{3.9 MB}\\
        \midrule
        \multirow{3}{*}{\rotatebox{90}{Breakfast}} & \xmark  & 55.6 & 52.3 & 47.3 / 42.1 / 31.4 & 91 MB\\
         & Random & 52.3 & 39.9 & 41.2 / 36.5 / 24.1 & 44 MB\\
         & \cellcolor{gray!30}{Ours} & \cellcolor{gray!30}{55.5} & \cellcolor{gray!30}{45.6} & \cellcolor{gray!30}{46.7 / 41.1 / 28.7} & \cellcolor{gray!30}{44 MB}\\
        \bottomrule
        \end{tabular}
\caption{Effectiveness of the sequence sampling strategies on three TAS benchmarks. Our proposed sampling outperforms random while retaining comparable performances to the case where no sequence sub-sampling is performed.}\label{tab:sub}
\end{table}

\begin{table}[!t]
\centering
\begin{tabular}{@{}c@{\hskip 6pt}c@{\hskip 6pt}c@{\hskip 6pt}c@{\hskip 6pt}r@{\hskip 6pt}c@{}}
\toprule
$\gamma$  & Acc & Edit & F1@\{10, 25, 50\} & Storage & Ratio(\%) \\ \midrule
0.1 & 45.0 & 44.9 & 40.5 / 35.6 / 23.2 & 1.3 MB & 0.53 \\
0.2 & 53.1 & 49.1 & 50.4 / 44.6 / 30.6 & 1.9 MB & 0.78 \\
0.3 & 56.9 & 52.3 & 56.8 / 52.6 / 36.7 & 2.4 MB & 0.98 \\
0.4 & 73.6 & \textbf{72.9} & 77.1 / 74.2 / \textbf{63.1} & 2.9 MB & 1.18 \\\rowcolor{gray!30}
0.5 & \textbf{75.2} & 71.9 & \textbf{78.3} / \textbf{74.6} / 62.7 & 3.6 MB & 1.47 \\ \midrule
1 & 76.3 & 74.8 & 80.0 / 78.0 / 61.7 & 7.2 MB & 2.94\\\bottomrule
\end{tabular}%
\caption{Sequence sampling ratio ($\gamma$) effects on GTEA. With only 0.5, we can achieve comparable performances to the full $\gamma=1$. %
}\label{tab:gamma}
\end{table}
\subsection{Ablation and Hyper-parameter Study}
\begin{figure}[htb]
        \centering
        \includegraphics[width=0.35\textwidth]{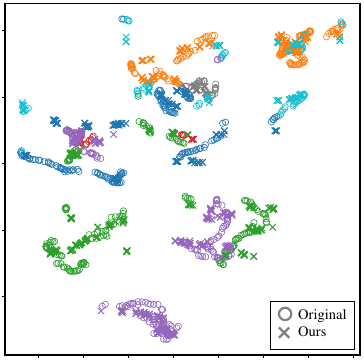} 
        \caption{T-SNE visualization of original and decoded video features. Different colors indicate different action classes. The visualization shows that our generated features are well-aligned with original features. Best viewed when zoomed in.}\label{fig:viz}
\end{figure}

\textbf{Sequence Sampling Strategy.} To evaluate the effectiveness of our proposed diversity-based sequence sampling technique, we compare it against random sampling and report the results in~\cref{tab:sub}. For all datasets, the default sampling ratio $\gamma$ is set to 0.5.
We first observe that, with a sampling ratio of 0.5, effectively reducing the number of samples by half, the segmentation performance is not significantly affected, highlighting sample redundancy in the video datasets.
On the other hand, our strategy consistently outperforms the random sampling across all metrics. Specifically, on the 50Salads dataset, there is a 2.5\% gap in the frame-wise accuracy (74.4\% vs. 71.9\%). The consistent performance gain over the counterpart underscores that, when constrained by a sequence budget, prioritizing the incorporation of diverse action sequences enhances the model's generalization capability more effectively. 

\begin{table}[!t]
\centering
\resizebox{\linewidth}{!}{
\begin{tabular}{@{}c@{\hskip 6pt}c@{\hskip 6pt}c@{\hskip 6pt}c@{\hskip 6pt}r@{\hskip 6pt}c@{}}
\toprule
\multicolumn{1}{c}{IPS ($K$)} & Acc & Edit & F1@\{10, 25, 50\} & Storage & Ratio(\%) \\ \midrule
\textcolor{black}{Mean} & \textcolor{black}{47.6} & \textcolor{black}{31.8} & \textcolor{black}{27.8} / \textcolor{black}{23.3} / \textcolor{black}{15.6} & \textcolor{black}{96 MB} & \textcolor{black}{0.34} \\\midrule
1 & 52.1 & 32.2 & 28.1 / 23.7 / 16.0 & 11 MB & 0.04 \\
2 & 52.7 & 38.4 & 34.9 / 30.1 / 21.1 & 22 MB & 0.08 \\
4 & 52.4 & 45.9 & 40.7 / 35.8 / 26.0 & 45 MB & 0.15 \\\rowcolor{gray!30}
8 & \textbf{55.6} & \textbf{52.3} & 47.3 / \textbf{42.1} / \textbf{31.4} & 91 MB & 0.31 \\
16 & 54.2 & 51.2 & \textbf{47.4} / 41.8 / 31.0 & 182 MB & 0.62 \\\midrule
$\dagger$ & 54.1 & 53.3 & 49.8 / 44.3 / 33.1 & 3.4 GB & 12.0 \\ \bottomrule
\end{tabular}}
\caption{Effect of the number of instance per segment ($K$) on Breakfast dataset without sequence sampling imposed. The ratio denotes the relative storage size of each setup compared to the original full dataset size. $\dagger$ indicates the setup in which latent codes are optimized on a per-frame basis. }\label{tab:ips}
\end{table}

\begin{table*}[!t]
\centering
\begin{tabular}{lc@{\hskip 6pt}c@{\hskip 6pt}c@{\hskip 6pt}c@{\hskip 6pt}c@{\hskip 6pt}c@{\hskip 6pt}c@{\hskip 6pt}c@{\hskip 6pt}c@{\hskip 6pt}c@{\hskip 6pt}}
\toprule
 \multirow{2}{*}{} & \multicolumn{4}{c}{MSTCN~\cite{farha2019ms}} & \multicolumn{4}{c}{ASFormer~\cite{yi2021asformer}} \\ \cmidrule(lr){2-5} \cmidrule(r){6-9}
 & Acc & Edit & \multicolumn{1}{c}{F1@\{10, 25, 50\}} & Avg & Acc & Edit & \multicolumn{1}{c}{F1@\{10, 25, 50\}}& Avg \\ \midrule
Mean~\cite{alssum2023just} & 18.4 & 14.1 & 14.4 / 13.0 / 9.8 & 13.9 & 12.1 & 10.4 & 10.7 / 9.7 / 7.9 & 10.2\\
TCA~\cite{ding2024coherent} & 31.4 & 25.0 & 25.5 / 22.9 / 17.4 & 28.3 & 36.0 & 31.9 & 32.4 / 29.3 / 22.8 & 30.5\\
\rowcolor{gray!30}
Ours & \textbf{38.2} & \textbf{31.6} & \textbf{32.8} / \textbf{29.7} / \textbf{22.8} &\textbf{31.0}&  \textbf{46.7} & \textbf{41.1} &  \textbf{41.7} / \textbf{38.3} / \textbf{30.8} & \textbf{39.7}\\\midrule
{Original} & {44.4} & {40.2} & {40.8 / 36.9 / 28.8 }& {38.2} & {51.7} & {46.2}& {47.2 / 43.4 / 34.7}& {44.6}\\
\bottomrule
\end{tabular}%
\caption{Performance comparison on the Breakfast dataset under the 10-task incremental setup following~\cite{ding2024coherent}. {``Avg'' indicates the averaged performance on all metrics}. Our approach consistently surpasses the counterparts, with both MSTCN and ASFormer backbones. }%
\label{tab:itas}
\end{table*}
\noindent \textbf{Sequence Sampling Ratio $\gamma$.} We further evaluate segmentation performances on the GTEA dataset using various sequence sampling ratios $\gamma$, as shown in~\cref{tab:gamma}. As $\gamma$ increases, a greater number of sequences are used to train the segmentation model, leading to a clear trend of improvement on all segmentation metrics. Notably, there is a substantial performance boost when $\gamma$ increases from 0.3 to 0.4, with a 16.7\% improvement in Acc and 20.6\% in the Edit score. Given the small scale of the GTEA dataset, a sampling ratio of 0.5 provides sufficient diversity in sampled sequences to effectively represent the dataset.

\noindent \textbf{Instances per Segment ($K$).} We next examine how the number of instances per segment ($K$) impacts the segmentation performance and storage. \Cref{tab:ips} presents the segmentation performance without imposing the sequence sampling. Across various $K$ values, our approach consistently outperforms the ``Mean'' baseline. Notably, even with $K=1$, requiring only $1/8$ of the storage (11 MB vs. 96 MB), a performance margin is achieved over ``Mean''. In both cases, each segment is condensed into a single representation, but with different dimensions. The ``mean'' approach retains the original feature dimension $D=2048$
while $K=1$ maps the segment into one latent code with dimension $d=256$.
This improvement demonstrates our method's effectiveness under extreme storage constraints.

Frame-wise Acc shows minimal sensitivity to $K$, with performance remaining relatively stable and a maximum variation of 3.5\% across different $K$ values. However, segmental metrics, such as Edit and F1 scores, demonstrate a clear upward trend as $K$ increases. For example, with $K=1$, where each segment is condensed into a single latent code, the Edit score is 32.2\%. Increasing $K$ to 8 boosts the score to 52.3\%, highlighting that a finer-grained condensation improves representation. This is because higher $K$ values allow each segment to be represented by multiple latent codes rather than a single, highly compressed one, creating a more detailed representation that better approximates the original data. 
However, performance plateaus at $K=16$, with no further improvement when condensation is conducted on a per-frame basis (denoted by $\dagger$). This suggests that the performance may be constrained by the generative model's expressiveness.
The storage size, as expected, increases linearly with $K$.
With $K=8$, the compression ratio for Breakfast is 0.31\%,  providing a good balance between storage efficiency and performance. 

\noindent \textbf{Visualization.} We plot both the original and decoded frame features using T-SNE~\cite{van2008visualizing} for a sample video sequence from the GTEA dataset in~\cref{fig:viz}. As shown, our network inversion approach effectively restores features that closely approximate the original, using the optimized latent codes. More visualizations are available in the Supplementary.

\section{Incremental Action Segmentation}\label{sec:itas}
One promising application of dataset condensation is in continual learning as it effectively alleviates the storage burden associated with replay data. 

We integrate our approach into the incremental temporal action segmentation (iTAS) framework recently proposed by~\cite{ding2024coherent}. In this setup, iTAS trains the segmentation model incrementally on different activity videos, with each stage focused on training with videos from a single activity. %
Each activity is treated as consisting of disjoint action classes, distinct from those of other activities. In their training, they assume that a small reservoir of samples from previous activities is considered available for the model to revisit, a process known as data replay. Our framework is applied to condense the replay data, the process is identical but on a per-activity basis. 

Following~\cite{ding2024coherent}, we conduct the experiment on the Breakfast dataset using the 10-task incremental setup, where each task corresponds to a single activity. Specifically, we train the TCA model for 2.5K epochs same as~\cite{ding2024coherent} and optimize the latent codes for the 10K iterations. To ensure the most efficient storage, we select the number of instances per segment as $K=1$. This choice aligns with~\cite{ding2024coherent}, where they also sample a single random latent code for each segment decoding.  
The results are summarized in~\cref{tab:itas}. With our approach applied, we achieve a 6.8\% increase in the final frame-wise accuracy with MSTCN~\cite{farha2019ms} and a 10.7\% increase with ASFormer~\cite{yi2021asformer}. Furthermore, our approach also significantly improves all segmental metrics by a margin larger than 10\% with ASFormer.

\section{Conclusion}\label{sec:conclusion}
This work introduces the first study on dataset condensation of temporal action segmentation. We propose a novel condensation framework to tackle the unique challenges of handling long procedural videos. 
It first condenses video segments into compact latent codes through generative network inversion from both temporal and channel perspectives. A diverse sequence sampling is further proposed to reduce the video-wise redundancy. Results on common benchmarks and with different backbones show our framework significantly reduces storage requirements, while preserving performance comparable to the original. This framework offers a practical solution for effectively condensing TAS datasets.

\vspace{1em}
\noindent \textbf{Acknowledgment} This research / project is supported by the Ministry of Education, Singapore, under the Academic Research Fund Tier 1 (FY2022).

\clearpage
{
    \small
    \bibliographystyle{ieeenat_fullname}
    \bibliography{main}

\begin{thebibliography}{46}
\providecommand{\natexlab}[1]{#1}
\providecommand{\url}[1]{\texttt{#1}}
\expandafter\ifx\csname urlstyle\endcsname\relax
  \providecommand{\doi}[1]{doi: #1}\else
  \providecommand{\doi}{doi: \begingroup \urlstyle{rm}\Url}\fi

\bibitem[com()]{commoncrawl}
Common crawl.
\newblock \url{https://commoncrawl.org/about/}.

\bibitem[Alssum et~al.(2023)Alssum, Alc{\'a}zar, Ramazanova, Zhao, and Ghanem]{alssum2023just}
Lama Alssum, Juan~Le{\'o}n Alc{\'a}zar, Merey Ramazanova, Chen Zhao, and Bernard Ghanem.
\newblock Just a glimpse: Rethinking temporal information for video continual learning.
\newblock In \emph{CVPRW}, 2023.

\bibitem[Carreira and Zisserman(2017)]{carreira2017quo}
Joao Carreira and Andrew Zisserman.
\newblock Quo vadis, action recognition? a new model and the kinetics dataset.
\newblock In \emph{CVPR}, 2017.

\bibitem[Cazenavette et~al.(2022)Cazenavette, Wang, Torralba, Efros, and Zhu]{cazenavette2022dataset}
George Cazenavette, Tongzhou Wang, Antonio Torralba, Alexei~A Efros, and Jun-Yan Zhu.
\newblock Dataset distillation by matching training trajectories.
\newblock In \emph{CVPR}, 2022.

\bibitem[Cazenavette et~al.(2023)Cazenavette, Wang, Torralba, Efros, and Zhu]{cazenavette2023generalizing}
George Cazenavette, Tongzhou Wang, Antonio Torralba, Alexei~A Efros, and Jun-Yan Zhu.
\newblock Generalizing dataset distillation via deep generative prior.
\newblock In \emph{CVPR}, 2023.

\bibitem[Cui et~al.(2023)Cui, Wang, Si, and Hsieh]{cui2023scaling}
Justin Cui, Ruochen Wang, Si Si, and Cho-Jui Hsieh.
\newblock Scaling up dataset distillation to imagenet-1k with constant memory.
\newblock In \emph{ICML}, 2023.

\bibitem[Deng et~al.(2009)Deng, Dong, Socher, Li, Li, and Fei-Fei]{deng2009imagenet}
Jia Deng, Wei Dong, Richard Socher, Li-Jia Li, Kai Li, and Li Fei-Fei.
\newblock Imagenet: A large-scale hierarchical image database.
\newblock In \emph{CVPR}, 2009.

\bibitem[Ding and Yao(2022{\natexlab{a}})]{ding2022leveraging}
Guodong Ding and Angela Yao.
\newblock Leveraging action affinity and continuity for semi-supervised temporal action segmentation.
\newblock In \emph{ECCV}, 2022{\natexlab{a}}.

\bibitem[Ding and Yao(2022{\natexlab{b}})]{ding2022temporal}
Guodong Ding and Angela Yao.
\newblock Temporal action segmentation with high-level complex activity labels.
\newblock \emph{IEEE TMM}, 2022{\natexlab{b}}.

\bibitem[Ding et~al.(2023)Ding, Sener, and Yao]{ding2023temporal}
Guodong Ding, Fadime Sener, and Angela Yao.
\newblock Temporal action segmentation: An analysis of modern techniques.
\newblock \emph{IEEE TPAMI}, 2023.

\bibitem[Ding et~al.(2024)Ding, Golong, and Yao]{ding2024coherent}
Guodong Ding, Hans Golong, and Angela Yao.
\newblock Coherent temporal synthesis for incremental action segmentation.
\newblock In \emph{CVPR}, 2024.

\bibitem[Farha and Gall(2019)]{farha2019ms}
Yazan~Abu Farha and Jurgen Gall.
\newblock Ms-tcn: Multi-stage temporal convolutional network for action segmentation.
\newblock In \emph{CVPR}, 2019.

\bibitem[Fathi et~al.(2011)Fathi, Ren, and Rehg]{fathi2011learning}
Alireza Fathi, Xiaofeng Ren, and James~M Rehg.
\newblock Learning to recognize objects in egocentric activities.
\newblock In \emph{CVPR}, 2011.

\bibitem[Fayyaz and Gall(2020)]{fayyaz2020sct}
Mohsen Fayyaz and Jurgen Gall.
\newblock Sct: Set constrained temporal transformer for set supervised action segmentation.
\newblock In \emph{CVPR}, 2020.

\bibitem[Kim et~al.(2022)Kim, Kim, Oh, Yun, Song, Jeong, Ha, and Song]{kim2022dataset}
Jang-Hyun Kim, Jinuk Kim, Seong~Joon Oh, Sangdoo Yun, Hwanjun Song, Joonhyun Jeong, Jung-Woo Ha, and Hyun~Oh Song.
\newblock Dataset condensation via efficient synthetic-data parameterization.
\newblock In \emph{ICML}, 2022.

\bibitem[Kuehne et~al.(2014)Kuehne, Arslan, and Serre]{kuehne2014language}
Hilde Kuehne, Ali Arslan, and Thomas Serre.
\newblock The language of actions: Recovering the syntax and semantics of goal-directed human activities.
\newblock In \emph{CVPR}, 2014.

\bibitem[Kuehne et~al.(2017)Kuehne, Richard, and Gall]{kuehne2017weakly}
Hilde Kuehne, Alexander Richard, and Juergen Gall.
\newblock Weakly supervised learning of actions from transcripts.
\newblock \emph{Computer Vision and Image Understanding}, 163:\penalty0 78--89, 2017.

\bibitem[Kukleva et~al.(2019)Kukleva, Kuehne, Sener, and Gall]{kukleva2019unsupervised}
Anna Kukleva, Hilde Kuehne, Fadime Sener, and Jurgen Gall.
\newblock Unsupervised learning of action classes with continuous temporal embedding.
\newblock In \emph{CVPR}, 2019.

\bibitem[Li and Todorovic(2020)]{li2020set}
Jun Li and Sinisa Todorovic.
\newblock Set-constrained viterbi for set-supervised action segmentation.
\newblock In \emph{CVPR}, 2020.

\bibitem[Li et~al.(2021)Li, Abu~Farha, and Gall]{li2021temporal}
Zhe Li, Yazan Abu~Farha, and Jurgen Gall.
\newblock Temporal action segmentation from timestamp supervision.
\newblock In \emph{CVPR}, 2021.

\bibitem[Liu et~al.(2023)Liu, Li, Dinh, Jiang, Shah, and Xu]{liu2023diffusion}
Daochang Liu, Qiyue Li, Anh-Dung Dinh, Tingting Jiang, Mubarak Shah, and Chang Xu.
\newblock Diffusion action segmentation.
\newblock In \emph{ICCV}, 2023.

\bibitem[Liu et~al.(2022)Liu, Wang, Yang, Ye, and Wang]{liu2022dataset}
Songhua Liu, Kai Wang, Xingyi Yang, Jingwen Ye, and Xinchao Wang.
\newblock Dataset distillation via factorization.
\newblock \emph{NeurIPS}, 35, 2022.

\bibitem[Loo et~al.(2022)Loo, Hasani, Amini, and Rus]{loo2022efficient}
Noel Loo, Ramin Hasani, Alexander Amini, and Daniela Rus.
\newblock Efficient dataset distillation using random feature approximation.
\newblock \emph{NeurIPS}, 2022.

\bibitem[Nguyen et~al.(2020)Nguyen, Chen, and Lee]{nguyen2020dataset}
Timothy Nguyen, Zhourong Chen, and Jaehoon Lee.
\newblock Dataset meta-learning from kernel ridge-regression.
\newblock \emph{arXiv preprint arXiv:2011.00050}, 2020.

\bibitem[Nguyen et~al.(2021)Nguyen, Novak, Xiao, and Lee]{nguyen2021dataset}
Timothy Nguyen, Roman Novak, Lechao Xiao, and Jaehoon Lee.
\newblock Dataset distillation with infinitely wide convolutional networks.
\newblock \emph{NeurIPS}, 2021.

\bibitem[Qi et~al.(2017)Qi, Yi, Su, and Guibas]{qi2017pointnet++}
Charles~Ruizhongtai Qi, Li Yi, Hao Su, and Leonidas~J Guibas.
\newblock Pointnet++: Deep hierarchical feature learning on point sets in a metric space.
\newblock \emph{NeurIPS}, 2017.

\bibitem[Rahaman et~al.(2022)Rahaman, Singhania, Thiery, and Yao]{rahaman2022generalized}
Rahul Rahaman, Dipika Singhania, Alexandre Thiery, and Angela Yao.
\newblock A generalized and robust framework for timestamp supervision in temporal action segmentation.
\newblock In \emph{ECCV}, 2022.

\bibitem[Richard et~al.(2018)Richard, Kuehne, and Gall]{richard2018action}
Alexander Richard, Hilde Kuehne, and Juergen Gall.
\newblock Action sets: Weakly supervised action segmentation without ordering constraints.
\newblock In \emph{CVPR}, 2018.

\bibitem[Sarfraz et~al.(2021)Sarfraz, Murray, Sharma, Diba, Van~Gool, and Stiefelhagen]{sarfraz2021temporally}
Saquib Sarfraz, Naila Murray, Vivek Sharma, Ali Diba, Luc Van~Gool, and Rainer Stiefelhagen.
\newblock Temporally-weighted hierarchical clustering for unsupervised action segmentation.
\newblock In \emph{CVPR}, 2021.

\bibitem[Sener and Yao(2018)]{sener2018unsupervised}
Fadime Sener and Angela Yao.
\newblock Unsupervised learning and segmentation of complex activities from video.
\newblock In \emph{CVPR}, 2018.

\bibitem[Shen and Elhamifar(2024)]{shen2024progress}
Yuhan Shen and Ehsan Elhamifar.
\newblock Progress-aware online action segmentation for egocentric procedural task videos.
\newblock In \emph{CVPR}, 2024.

\bibitem[Singhania et~al.(2022)Singhania, Rahaman, and Yao]{singhania2022iterative}
Dipika Singhania, Rahul Rahaman, and Angela Yao.
\newblock Iterative contrast-classify for semi-supervised temporal action segmentation.
\newblock In \emph{AAAI}, 2022.

\bibitem[Stein and McKenna(2013)]{stein2013combining}
Sebastian Stein and Stephen~J McKenna.
\newblock Combining embedded accelerometers with computer vision for recognizing food preparation activities.
\newblock In \emph{UbiComp}, 2013.

\bibitem[Van~der Maaten and Hinton(2008)]{van2008visualizing}
Laurens Van~der Maaten and Geoffrey Hinton.
\newblock Visualizing data using t-sne.
\newblock \emph{Journal of machine learning research}, 9\penalty0 (11), 2008.

\bibitem[Wang et~al.(2022)Wang, Zhao, Peng, Zhu, Yang, Wang, Huang, Bilen, Wang, and You]{wang2022cafe}
Kai Wang, Bo Zhao, Xiangyu Peng, Zheng Zhu, Shuo Yang, Shuo Wang, Guan Huang, Hakan Bilen, Xinchao Wang, and Yang You.
\newblock Cafe: Learning to condense dataset by aligning features.
\newblock In \emph{CVPR}, 2022.

\bibitem[Wang et~al.(2018)Wang, Zhu, Torralba, and Efros]{wang2018dataset}
Tongzhou Wang, Jun-Yan Zhu, Antonio Torralba, and Alexei~A Efros.
\newblock Dataset distillation.
\newblock \emph{arXiv preprint arXiv:1811.10959}, 2018.

\bibitem[Wang et~al.(2024)Wang, Xu, Lu, and Li]{wang2024dancing}
Ziyu Wang, Yue Xu, Cewu Lu, and Yong-Lu Li.
\newblock Dancing with still images: Video distillation via static-dynamic disentanglement.
\newblock In \emph{CVPR}, 2024.

\bibitem[Welling(2009)]{welling2009herding}
Max Welling.
\newblock Herding dynamical weights to learn.
\newblock In \emph{ICML}, 2009.

\bibitem[Yi et~al.(2021)Yi, Wen, and Jiang]{yi2021asformer}
Fangqiu Yi, Hongyu Wen, and Tingting Jiang.
\newblock Asformer: Transformer for action segmentation.
\newblock In \emph{BMVC}, 2021.

\bibitem[Yu et~al.(2023)Yu, Liu, and Wang]{yu2023dataset}
Ruonan Yu, Songhua Liu, and Xinchao Wang.
\newblock Dataset distillation: A comprehensive review.
\newblock \emph{IEEE Transactions on Pattern Analysis and Machine Intelligence}, 2023.

\bibitem[Zhang et~al.(2023)Zhang, Wang, Xue, Yan, Zhang, Bai, and Shou]{zhang2023dataset}
David~Junhao Zhang, Heng Wang, Chuhui Xue, Rui Yan, Wenqing Zhang, Song Bai, and Mike~Zheng Shou.
\newblock Dataset condensation via generative model.
\newblock \emph{arXiv preprint arXiv:2309.07698}, 2023.

\bibitem[Zhao and Bilen(2021)]{zhao2021dataset}
Bo Zhao and Hakan Bilen.
\newblock Dataset condensation with differentiable siamese augmentation.
\newblock In \emph{ICML}, 2021.

\bibitem[Zhao and Bilen(2023)]{zhao2023dataset}
Bo Zhao and Hakan Bilen.
\newblock Dataset condensation with distribution matching.
\newblock In \emph{WACV}, 2023.

\bibitem[Zhao et~al.(2020)Zhao, Mopuri, and Bilen]{zhao2020dataset}
Bo Zhao, Konda~Reddy Mopuri, and Hakan Bilen.
\newblock Dataset condensation with gradient matching.
\newblock \emph{arXiv preprint arXiv:2006.05929}, 2020.

\bibitem[Zhong et~al.(2024)Zhong, Ding, and Yao]{zhong2024onlinetas}
Qing Zhong, Guodong Ding, and Angela Yao.
\newblock Onlinetas: An online baseline for temporal action segmentation.
\newblock \emph{NeurIPS}, 2024.

\bibitem[Zhou et~al.(2022)Zhou, Nezhadarya, and Ba]{zhou2022dataset}
Yongchao Zhou, Ehsan Nezhadarya, and Jimmy Ba.
\newblock Dataset distillation using neural feature regression.
\newblock \emph{NeurIPS}, 2022.

\end{thebibliography}
}

\end{document}